\documentclass[10pt,twocolumn,letterpaper]{article}

\usepackage{cvpr}              % To produce the CAMERA-READY version
\usepackage{graphicx}
\usepackage{amsmath}
\usepackage{amssymb}
\usepackage{booktabs}
\usepackage{algorithm}
\usepackage{algorithmicx}
\usepackage{amsfonts}
\usepackage{amsmath}
\usepackage{bbm}
\usepackage{booktabs}
\usepackage{multirow}
\usepackage{makecell}
\usepackage{threeparttable}
\algnewcommand\algorithmicinput{\textbf{Input:}}
\algnewcommand\For{\textbf{For:}}
\algnewcommand\EndFor{\textbf{Endfor:}}
\algnewcommand\INPUT{\item[\algorithmicinput]}
\algnewcommand\algorithmicoutput{\textbf{Output:}}
\algnewcommand\OUTPUT{\item[\algorithmicoutput]}

\usepackage[pagebackref,breaklinks,colorlinks]{hyperref}

\usepackage[capitalize]{cleveref}
\crefname{section}{Sec.}{Secs.}
\Crefname{section}{Section}{Sections}
\Crefname{table}{Table}{Tables}
\crefname{table}{Tab.}{Tabs.}

\def\cvprPaperID{11971} % *** Enter the CVPR Paper ID here
\def\confName{CVPR}
\def\confYear{2022}

\begin{document}

\title{Pruning-aware Sparse Regularization for Network Pruning}

\author{
Nanfei Jiang$^{1,2,*}$, Xu Zhao$^{1,}$\thanks{Equal contribution.}\ , Chaoyang Zhao$^{1}$, Yongqi An$^{1,2}$, Ming Tang$^{1,2}$, Jinqiao Wang$^{1,2}$\\
\\
1. National Laboratory of Pattern Recognition, Institute of Automation,\\ Chinese Academy of Sciences,Beijing, 100190, China.
\\
2. University of Chinese Academy of Sciences, Beijing, 100049, China \\
{\tt\small \{nanfei.jiang, xu.zhao, chaoyang.zhao, yongqi.an, tangm,  jqwang\}@nlpr.ia.ac.cn}
}
\maketitle

%%%%%%%%% ABSTRACT
\begin{abstract}
Structural neural network pruning aims to remove the redundant channels in the deep convolutional neural networks (CNNs) by pruning the filters of less importance to the final output accuracy. To reduce the degradation of performance after pruning, many methods utilize the loss with sparse regularization to produce structured sparsity. In this paper, we analyze these sparsity-training-based methods and find that the regularization of unpruned channels is unnecessary. Moreover, it restricts the network's capacity, which leads to under-fitting. To solve this problem, we propose a novel pruning method, named MaskSparsity, with pruning-aware sparse regularization. MaskSparsity imposes the fine-grained sparse regularization on the specific filters selected by a pruning mask, rather than all the filters of the model. Before the fine-grained sparse regularization of MaskSparity, we can use many methods to get the pruning mask, such as running the global sparse regularization. MaskSparsity achieves 63.03\%-FLOPs reduction on ResNet-110 by removing 60.34\% of the parameters, with no top-1 accuracy loss on CIFAR-10. On ILSVRC-2012, MaskSparsity reduces more than 51.07\% FLOPs on ResNet-50, with only a loss of 0.76\% in the top-1 accuracy. 

The code is released at \url{https://github.com/CASIA-IVA-Lab/MaskSparsity}. Moreover, we have integrated the code of  MaskSparity into a PyTorch pruning toolkit, EasyPruner, at \url{https://gitee.com/casia_iva_engineer/easypruner}.

\begin{figure}[h!]
    \begin{centering}
        
        \includegraphics[width=0.9\linewidth]{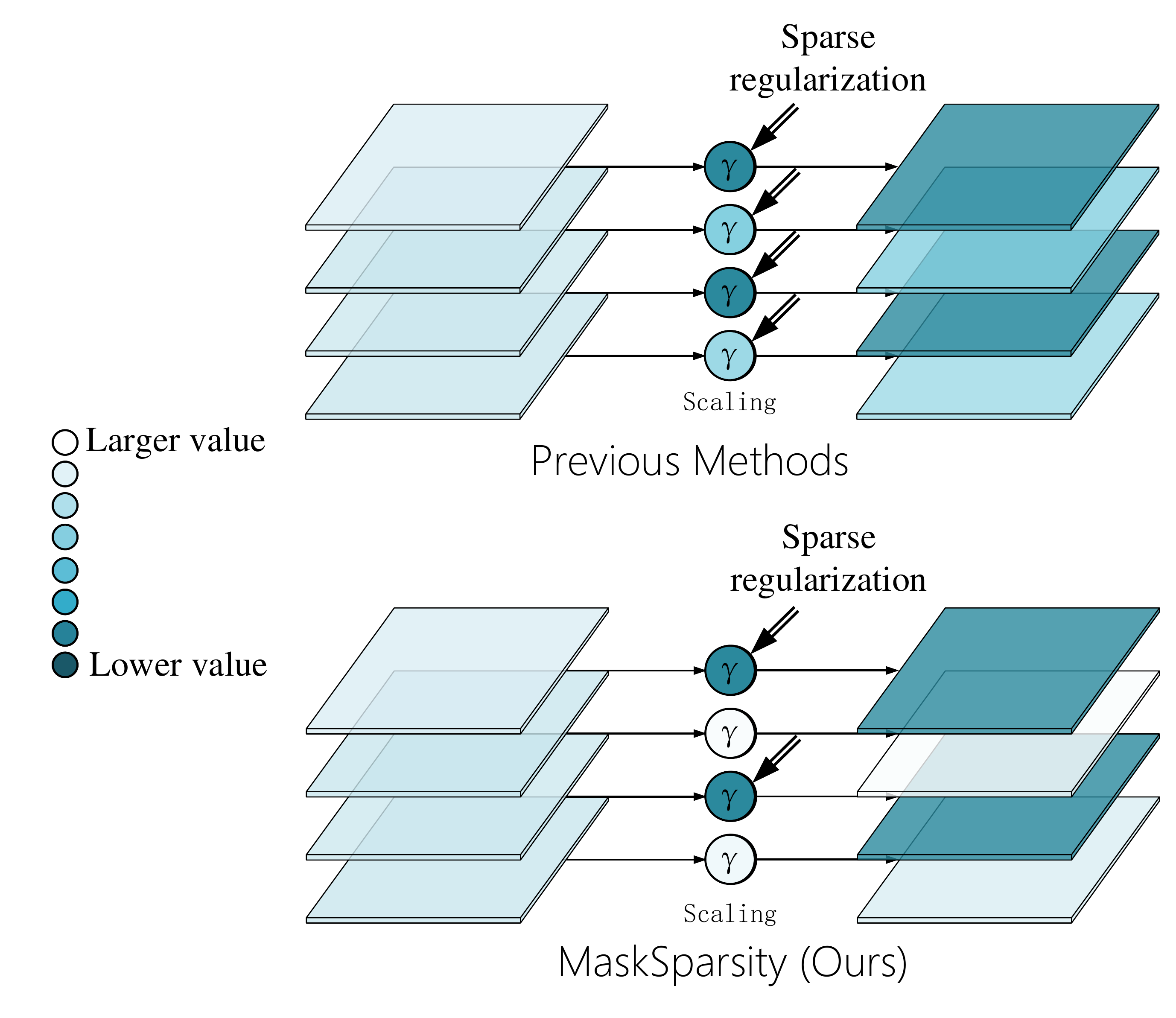}
        \vspace{-0.2cm}
        \caption{Visual comparison between the previous sparsity-training-based methods and the proposed MaskSparsity method. MaskSparity apply the sparse regularization only on the scaling factors of less-important channels.}
        \label{MAskSparsity}
        \par\end{centering} 
            \vspace{-0.4cm}
\end{figure}
\end{abstract}

\begin{figure*}[h!]
    \begin{centering}
        \includegraphics[width=0.9\linewidth]{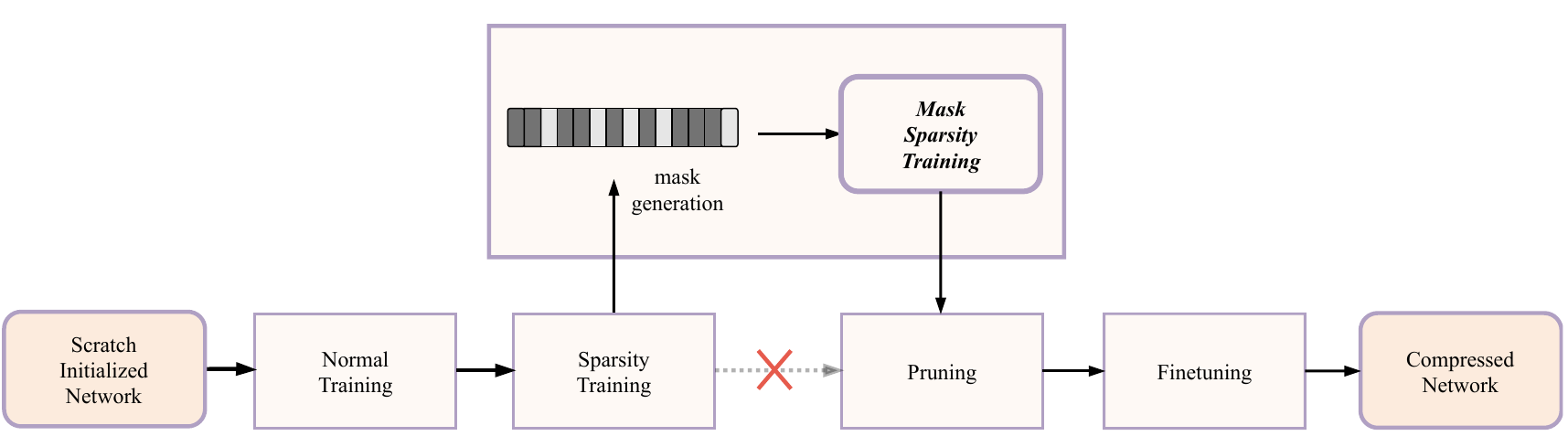}
        \vspace{-0.3cm}
        \caption{The pipeline of the proposed MaskSparsity method.}
        \vspace{-0.3cm}
        \label{MAskSparsityPipeline}
        \par\end{centering} 
\end{figure*}

%%%%%%%%% BODY TEXT
\section{Introduction}
\label{sec:intro}

\noindent Convolutional Neural Networks (CNNs) have demonstrated a great success on a variety of computer vision tasks, like image classification \cite{russakovsky2015imagenet}, detection \cite{lin2014microsoft}, and semantic segmentation \cite{Cordts2016Cityscapes}.  However, the increasing depth and width of the CNNs also lead to higher computing resources demands and excessive memory footprint requirements. Typically, the widely used ResNet models \cite{he2016deep} have millions of parameters, requiring billions of float point operations (FLOPs), making it a great challenge to deploy most state-of-the-art CNNs on edge devices. 
Network pruning is an effective way to compress and accelerate CNNs. It is attracting much attention from researchers. It can remove the parameters in the deep CNNs and reduce the required FLOPs and memory footprint while preserving the performance. 

A typical scheme of network pruning consists of three stages: (1) training an over-parameterized model normally; (2) pruning the model under a certain criterion; and (3) fine-tuning the pruned model to reduce the degradation caused by pruning. Some of the existing network pruning methods apply a sparsity training stage after step (1). These methods apply sparse regularization on the filter weights of the convolution layers \cite{alvarez2016learning,wen2016learning} or scaling factors \cite{huang2018data,liu2017learning} of the batch normalization layers. After the sparsity training, the corresponding filter weights or scaling factors of unimportant channels are considered to be near zero. Then these channels could be safely pruned without affecting the output values of the corresponding layers too much. We call these methods sparsity-training-based methods. 

In the sparsity-training-based methods, to get an expected sparse rate of the model,  they adopt the global sparse regularization. However, in these methods,
the weights of important channels are regularized in the sparsity training stage, although they are preserved after the pruning.
It is generally regarded that proper regularization achieved a good result by avoiding over-fitting. However, the model will be under-fitted when the regularization coefficient is too large.  Since the weights of important channels are also regularized by the sparse regularization, the magnitude of these weights is usually decayed towards 0.  
This prevents the coverage to better local minima of the network in the sparsity training stage, which affects the final performance of the fine-tuned pruned network. 
Moreover, the sparse rate of the globally trained network is hard to control. 
This usually results in the inconsistency between the sparse mask and the pruning mask if we want to prune a model to a predefined FLOPS.

To address the problem mentioned above, we propose a novel sparsity-training-based channel pruning approach, MaskSparsity. 
Different from the previous sparsity-training-based methods which impose the regularization on all channels of each layer, MaskSparsity only imposes the regularization on the specific channels selected by the pruning mask, which indicates the unimportant channels, as shown in Figure \ref{MAskSparsity}. 
Through this mask, MaskSparsity can realize the strong correlation between pruning and regularization, and carry out a pruning-aware regularization. 
In other words, we only impose regularization on the channels to be pruned and prune the channels where regularization is applied. 
The perfect match between the sparse channels and the pruning channels allows us to minimize the impact of sparse regularization and maximize the accuracy of the pruned networks. 
  
Compared with the typical pruning methods which directly prune the unimportant channels, the MaskSparsity can gradually push the unimportant parameters towards zero in a long period of iterations during the sparsity training stage. This prevents the model from a dramatic change of structure or weight in the pruning stage. It is regarded that the dramatic change may result in a certain amount of information loss which is harmful to restoring the accuracy of the model in the fine-tuning stage.

To summarize, our main contributions are three-fold:

\begin{itemize}
\item We analyze the previous sparsity-training-based methods in the previous work, which simply impose L1 regularization on all channels of the model. We find out the over-regularization problem on important channels.
\item We propose MaskSparsity to solve these problems by more fine-grained sparsity training.
\item The extensive experiments on two benchmarks show the effectiveness and efficiency of MaskSparsity.
\end{itemize}

%------------------------------------------------------------------------
\section{Related Work}
We mainly focus on the structural pruning methods in this paper.
In this section, we first review the closely related works, \ie the sparsity-training-based structural pruning methods. After that, we list other structural pruning methods.

\subsection{Sparsity-training-based Pruning Methods}
To make the network adaptively converge to a sparse structure and alleviate the damage of the pruning process to the network's output, some sparsity-training-based pruning methods are proposed. There are mainly two categories of these methods according to the place where sparse regularization is applied. 

The first category of methods is the group-sparsity-based methods that apply spare regularization on the filter weights. Alvarez and Salzmann \cite{alvarez2016learning} proposed to use a group sparsity regularizer to determine the number of channels of each layer. Wen \etal \cite{wen2016learning} proposed a Structured Sparsity Learning (SSL) method to regularize the structure to obtain a hardware-friendly pruned structure.  Alvarez and Salzmann \cite{alvarez2017compression} added a low-rank regularizer to improve the pruning performance.  Li and Gu proposed the Hinge \cite{Hinge} by combining the filter pruning and low-rank decomposition into the group sparsity training framework.

The second category of methods is the indirect group-sparse methods, which apply the sparse regularization on the scaling factors of each layer. The representative method is NetSlim \cite{liu2017learning} method, which sparsely regularizes the scaling factors of BN layers to get the sparse structure and remove less important channels. Huang and Wang \cite{huang2018data} proposed to add a new scaling factor vector to each layer to apply the sparse regularization. Srinivas and Subramanya \cite{Srinivas2017TrainingSN} proposed to impose sparse constraint over each weight with additional gate variables and achieve high compression rates by pruning connections with zero gate values. Ye and You \cite{OT} proposed to prune channels with layer-dependent thresholds according to the different weight distribution of each layer. \cite{Dependency} develop the norm-based importance estimation by taking the dependency between the adjacent layers into consideration.

These methods apply the global sparse regularization on the network channels, which over-regularize the important channels. Our MaskSparsity method solves this problem and can improve the performance of sparsity-training-based methods to the new state-of-the-art.

\subsection{Non-sparsity-training-based Pruning Methods}
Recently, many non-sparsity-training-based pruning methods also show good performance. These methods usually evaluate the importance of each channel with a handcraft criterion first. After that, they directly prune the unimportant channels and finetune the network.
For instance, Li and Kadav \cite{L1} proposed to prune filters with smaller L1 norm values in a network. Based on the theory of Geometric Median (GM) \cite{GM}, He and Liu proposed FPGM\cite{FPGM} to prune the filters with the most replaceable contribution. Inspired by the discovery that the average rank of multiple feature maps generated by a single filter is always the same, Lin \etal  \cite{lin2020hrank} proposed to prune filters by exploring the High Rank of feature maps (HRank).  
In this paper, we compare the performance of the proposed MaskSparsity with these methods and show good pruning performance.

%------------------------------------------------------------------------

\section{Methodology}

\subsection{Notations}\label{Notations-and-background}
We assume that a convolutional neural network consists of multiple convolutional layers and each convolution layer is followed by a batch normalization (BN) \cite{ioffe2015batch} layer. For the $l$-th convolutional layer, we use $C_l$ and $N_l$ to represent the number of its input channels and output channels, and $k_l\times k_l$ represent the kernel size. 

We use $\mathcal{W}^{(l)} = \{ \mathcal{W}^{(l)}_{1,:,:,:}, \mathcal{W}^{(l)}_{2,:,:,:}, ..., \mathcal{W}^{(l)}_{N_l,:,:,:} \} \in \mathbb{R}^{ N_l \times C_l \times k_l \times k_l}$ to represent the filters of the  $l$-th convolutional layer. The input feature maps and the output feature maps to filters are denoted as $\mathcal{I}^{(l)} = \{ \mathbf{i}_1^{(l)}, \mathbf{i}_2^{(l)}, ..., \mathbf{i}_{C_l}^{(l)} \} \in \mathbb{R}^{ B \times  C_l \times h_l \times w_l}$ and $\mathcal{O}^{(l)} = \{ \mathbf{o}_1^{(l)}, \mathbf{o}_2^{(l)}, ..., \mathbf{o}_{N_l}^{(l)} \} \in \mathbb{R}^{ B \times N_l \times {h_l}^\prime \times {w_l}^\prime}$. Here, $h_l$, $w_l$, ${h_l}^\prime$ and ${w_l}^\prime$ are the heights and widths of the input and output feature maps respectively. $B$ is the batch size of the input images. The $i$-th channel feature map $\mathbf{o}_i^{(l)} \in \mathbb{R}^{B \times {h_l}^\prime \times {w_l}^\prime}$ is generated by $\mathcal{W}^{(l)}_{i,:,:,:} \in \mathbb{R}^{C_l \times k_l \times k_l}$ and $\mathbf{I}^{(l)} \in \mathbb{R}^{B \times C_l \times h_l \times w_l}$.

For the $i$-th channel of the $l$-th BN layer with mean ${\mu}^{(l)}_i$, standard deviation ${\sigma}^{(l)}_i$, learned scaling factor ${\gamma}^{(l)}_i$ and bias ${\beta}^{(l)}_i$, regardless of bias of the convolutional layer, we have  
\begin{equation}\label{eq-conv-bn}
    \mathbf{o}_i^{(l)} = (\mathcal{W}^{(l)}_{i,:,:,:} \otimes  \mathbf{I}^{(l)} - \mu^{(l)}_i)\frac{\gamma^{(l)}_i}{\sigma^{(l)}_i} + \beta^{(l)}_i \,.
\end{equation}

\subsection{Existing Sparsity-training-based Methods}\label{Notations-and-background}
Existing sparsity-training-based methods utilize sparse 
regularization loss to produce structured sparsity. Usually, the sparse regularization is either applied on the filter weights of convolutions or the channel scaling factors of BNs. 

\textbf{(a) Sparsity on filter weights.} When sparse regularization is applied on the filter weights of convolutions, the training objective function of this category of methods is shown in Equation \ref{eq-tra-sp-lasso}:
\begin{equation}\label{eq-tra-sp-lasso}
    L_{\text{Sparsity}}(\mathcal{I}^0, y, \mathcal{W}) = L(f(\mathcal{I}^0, \mathcal{W}), y) 
    + \lambda \cdot \sum_{l=1}^{L} \sum_{i=1}^{N_l}||\mathcal{W}^{(l)}_{i,:,:,:}||_g   \,
    ,
\end{equation}
where $( \mathcal{I}^0$ , $y )$  denote the training samples  and the labels, $\mathcal{W}$ denotes the trainable weights, the $L(f(\mathcal{I}^0, \mathcal{W}), y)$ is the objective function of normal training, $||\mathcal{W}^{(l)}_{i,:,:,:}||_g$ is a sparsity regularization penalty on the filter weights $\mathcal{W}$, here $||\cdot||_g$ is the group Lasso, $||\mathcal{W}^{(l)}_{i,:,:,:}||_g =\sqrt{ \sum^{C_l}  \sum^{k_1} \sum^{k_1}    \left(\mathcal{W}^{(l)}_{i,j,k_1,k_2}\right)^2 }$. $\lambda$ is the factor of controlling the strength of sparsity.

\textbf{(b) Sparsity on channel scaling factors.} When sparse regularization is applied on the channel scaling factors of the BN layer, the training objective function of this category of methods is shown in Equation \ref{eq-tra-sp}:  
\begin{equation}\label{eq-tra-sp}
    L_{\text{Sparsity}}(\mathcal{I}^0, y, \mathcal{W}) = L(f(\mathcal{I}^0, \mathcal{W}), y) + 
    \lambda \cdot \sum_{l=1}^{L} \sum_{i=1}^{N_l}|| \gamma ^{(l)}_{i}||_g   \,
    ,
\end{equation}
where $( \mathcal{I}^0$ , $y )$, $L(f(\mathcal{I}^0, \mathcal{W}), y)$, 
and $\lambda$ denote the same mean as above. 
$|| \gamma ^{(l)}_{i}||_g$ is a sparsity regularization
penalty on the scaling factors $\gamma$ of BN layers.
$||\cdot||_g$ is usually set as L1 regularization, and L2
regularization is available either as \cite{Han2015LearningBW,Tartaglione2018LearningSN}.

In this paper, we choose the sparsity regularization penalty on \textbf{the channel scaling factors} for further investigation.

As with the normal regularization methods, the received gradients of the scaling factors from the normal training loss $L$ and the sparsity regularization $||\cdot||_g$ are usually against each other during training. The former aims at improving the model performance on the training set.  The latter aims at restricting the range of the parameters and increasing the structure sparsity, which tends to increase the loss on the training set. The sparsity-training-based pruning method thinks that the unimportant convolutional channels are easily pushed to near 0 by $||\cdot||_g$, while the value of important channels is kept large by $||\cdot||_g$.

%The $||\cdot||_g$ loss tends to retain the capacity of the network by pushing the learnable parameters to 0. The unimportant channels' gradients from the normal training loss $L$ are usually small, making them being more likely to be pushed to near 0 by $||\cdot||_g$. For the important channels, the gradients from the $L$ are large enough to prevent the scaling factors from being degraded to 0 by $||\cdot||_g$. 

\subsection{The Over-regularization Problem of Existing Sparsity-training-based Methods }\label{Analysisoftraditionalsp}

\begin{figure}[h!]
%   \vspace{-0.4cm}
    \begin{centering}
        \includegraphics[width=0.4\textwidth]{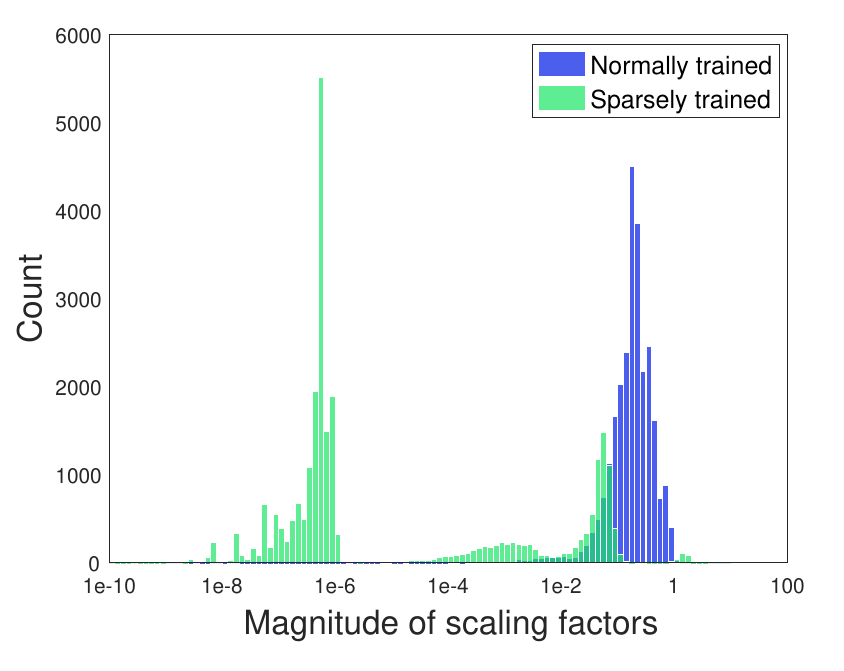} 
        \vspace{-0.4cm}
        \caption{Distribution of scaling factors of ResNet50 before and after global sparsity training.}
        \label{pbm}
        \vspace{-0.2cm}
        \par\end{centering}
    
\end{figure}

Figure \ref{pbm} shows the statistical results of two sets of scaling factors (absolute value) collected from the normally-trained and sparsely-trained ResNet-50  on ILSVRC-2012. In Figure \ref{pbm}, the purple histogram is the distribution of the normally-trained network, while the green histogram represents that of the sparse-trained network. Figure \ref{pbm} shows that the scaling factors of the normally-trained network form one peak and those of the sparsely-trained network form two peaks. This is consistent with the bimodal-distribution observation of OT \cite{ye2020channel}.

Obviously, the left scaling-factors peak of the sparsely-trained network represents the unimportant channels and the right peak represents the important channels. 
With Figure \ref{pbm}, we demonstrate the over-regularization problem of the existing sparsity-training-based methods, which are mentioned in the introduction.

It can be seen that the right peak of the sparsely-trained network moves to 0 obviously, compared with their location in the histogram of the normally-trained network. This is a common phenomenon of the model regularization methods, i.e. the regularization cause a smaller magnitude of the model parameters. A small regularization usually leaders to better generalization performance on the test set. However, a too large regularization leads to under-fitting. This is because the regularization limits the network's capacity. 

The modern CNNs are usually trained with \emph{weight decay}, 
which is widely regarded to be similar to L2 regularization, especially under the SGD optimizer \cite{goyal2017accurate}. This is usually tuned to a proper magnitude to get the best performance. Moreover, the sparse regularization is regarded to push a large partition of a channel to be near 0. This requires a large weight for the sparsity loss.
Therefore, we think the newly applied sparse regularization on the unpruned channels is over-regularization. It should be avoided.

%Since the network channels corresponding to the scaling factors of this peak are preserved after the network pruning, this phenomenon indicates that the unpruned important channels are over-regularized.  We think this over-regularization prevents the better coverage of the network in the sparsity training stage, which affects the final performance of the fine-tuned pruned network.

%Second, there are many in-between bars between the main bodies of the two peaks. This bad distribution of scaling factors challenges the threshold selecting. The channels of the in-between area are difficult to be classified as important or not. This challenges the selection of an optimal pruning threshold. With a sub-optimal threshold, some pruned channels may still be important for the network.

\subsection{Pruning-aware Sparse Regularization}\label{MaskSP}
Therefore, we propose a fine-grained sparsity training method that only applies the sparse regularization on the unimportant channels to keep a maximum representation ability of the important channels. 

The task of sparsity training consists of two sub-tasks implicitly. The first sub-task is identifying the unimportant channels. The second sub-task is pushing the filter weights or scaling factors of unimportant channels to 0 by the sparse regularization loss. Existing sparsity-training-based methods accomplish the two sub-tasks simultaneously in the sparsity training stage. We propose to decouple the two sub-tasks. By doing this, we can apply the fine-grained spare regularization, which only sparse out the unimportant channels.

Figure \ref{MAskSparsityPipeline} shows the training pipeline of the proposed MaskSparsity. We transform the sparsity training stage of the existing methods into two stages. The first stage is the sparsity training stage with global sparse regularization, which is aimed to get the indexes of the unimportant channels. The indexes are transformed into a binary pruning mask in previous methods. In our method, we use the mask to identify which channels to apply the sparse regularization in the second stage. To get the pruning mask, we directly threshold the scaling factors of the normally trained network. The details is shown in Equation \ref{maskgeneration}:
\begin{equation}
    \mathcal{M} = \{\mathbbm{1}(\gamma<\theta)|\gamma \in \Gamma\},
    \label{maskgeneration}
\end{equation}
where $\mathbbm{1}$ is the indicator function, $\Gamma$ is all the scaling factors of the network, and $\theta$ is the predefined pruning threshold of the pruning method. Actually, the pruning mask $\mathcal{M}$ consists of the unimportant-channel mask of each layers, \ie $\mathcal{M} = \{ \mathcal{M}^{(1)}, \mathcal{M}^{(2)}, ..., \mathcal{M}^{(L)} \} $, where $L$ is the layer count of the network. According to Equation \ref{maskgeneration}, the pruning masks $\mathcal{M}_i$ is a binary vector consisting of 0 and 1.

\begin{algorithm}[t]
    \caption{Algorithm Description of MaskSparsity}
    \label{alg:MaskSP}
    \begin{algorithmic}[1]
        \INPUT training data: $\{\mathcal{X},y\}$, pruning threshold $\theta$.
        
        \State \textbf{Initialize}: pretrained model parameter $\mathcal{W}=\{\mathcal{W} _{i}, 0\leq i \leq L\}$;  
        
        \For{$epoch=1$; $epoch \leq epoch_{max}$; $epoch++$}
        \State Update the model parameter $\mathcal{W}$ based on $\{\mathcal{X},y\}$ and, using the global sparse regularization as in Equation~\ref{eq-tra-sp};
        \EndFor
        
        \State Obtain the pruning mask $\mathcal{M}$ by thresholding $\gamma$ with $\theta$;
        
        \State  \textbf{Reinitialize}: pretrained model parameter $\mathcal{W}=\{\mathcal{W} _{i}, 0\leq i \leq L\}$;
        
        \For{$epoch=1$; $epoch \leq epoch_{max}$; $epoch++$}
        \State Update the model parameter $\mathcal{W}$ based on $\{\mathcal{X},y\}$ , the mask-guided sparse regularization as shown in Equation~\ref{eq-Mask-sp} with the mask $\mathcal{M}$;
        \EndFor
        \State Obtain the compact model $\mathcal{W} ^{*}$ from $\mathcal{W}$;
        \State Finetune the compact model $\mathcal{W} ^{*}$;
        \OUTPUT The compact model and its parameters $\mathcal{W} ^{*}$.
    \end{algorithmic} 
\end{algorithm}

As discussed above, the over-regularization of the important channels limits the network capacity. Therefore, in this paper, we design a fine-grained sparse training strategy to alleviate the damage of the sparse regularization loss on important channels.
Specifically, we propose to apply the sparse regularization only on the unimportant channels. Based on Equation \ref{eq-tra-sp}, we can describe our MaskSparsity method as the Equation \ref{eq-Mask-sp}:

\begin{equation}\label{eq-Mask-sp}
%   L_{\text{Sparsity}}(\mathcal{I}^0, y, \mathcal{W}) = L(f(\mathcal{I}^0, \mathcal{W}), y) + \lambda \cdot  \mathcal{M} \cdot   Reg({\gamma}) \,,
    L_{\text{Sparsity}}(\mathcal{I}^0, y, \mathcal{W}) = L(f(\mathcal{I}^0, \mathcal{W}), y)
    + \lambda \cdot \sum_{l=1}^{L} \sum_{i=1}^{N_l}  \mathcal{M}^{(l)}_{i}   || \gamma ^{(l)}_{i}||_g   \,,
\end{equation}
where $\mathcal{M}$ denotes the binary mask, indicating the unimportant channels of the whole network.  For the important channels, the values in the channel mask are 0. Therefore, these channels are not affected by the sparse regularization and are trained as normal.

\begin{table*}
    
    \caption{Evaluation results using ResNet-50 on ILSVRC-2012.} 
    %https://github.com/anonymous47823493/EagleEye
   \vspace{-0.6cm}
    \begin{center}
    \begin{threeparttable}
        \begin{small}
            
            \begin{tabular}{l|ccccccc}
                \toprule[1.5pt]
                 Method                             &\makecell[c]{Base  \\ Top-1(\%)}   &\makecell[c]{Base \\ Top-5(\%)}    &\makecell[c]{Pruned \\ Top-1(\%)}  &\makecell[c]{Pruned \\ Top-5(\%)}  & \makecell[c]{Top-1 $\downarrow$ \\ (\%)}  &   \makecell[c]{Top-5 $\downarrow$ \\ (\%)}    &   \makecell[c]{FLOPs $\downarrow$ \\ (\%)}    
                
                \\  \midrule[1pt] 
                
                   NS \cite{liu2017learning}          &  75.04    &    -      &   69.60   &    -      &    5.44   &  -        &   50.51   \\
                   OT  \cite{OT}               &  75.04    &    -      &   70.40   &    -      &    4.64   &  -        &   52.88   \\
                    SFP \cite{he2018soft}       &   76.15   &   92.87   &   74.61   &   92.06   &   1.54    &   0.81    &   41.8    \\
                GAL-0.5 \cite{GAL}              &   76.15   &   92.87   &   71.95   &   90.94   &   4.20    &   1.93    &   43.03   \\
                HRank \cite{HRank}              &   76.15   &   92.87   &   74.98   &   92.33   &   1.17    &   0.54    &   43.76   \\
                Hinge \cite{Hinge}             &     -     &    -      &   74.7    &    -     &     -     &       &  46.55           \\
                %&NISP \cite{yu2018nisp}            &   -       &   -       &   -       &   -       &   0.89    &   -       &   44.01   \\  
                %&\cite{singh2018stability}     &   -       &   92.65   &   -       &   92.2    &   -       &   0.45    &   44.45       \\  
                %&CFP \cite{singh2018leveraging}    &   75.3    &   92.2    &   73.4    &   91.4    &   1.9     &   0.8     &   49.6    \\
                %&Channel Pr \cite{he2017channel}&  -       &   92.2    &   -       &   90.8    &   -       &   1.4     &   50      \\
                %&SPP \cite{wang2017structured} &   -       &   91.2    &   -       &   90.4    &   -       &   0.8     &   50      \\
                HP \cite{xu2018hybrid}          &   76.01   &   92.93   &   74.87   &   92.43   &   1.14    &   0.50    &   50      \\
                %&ELR \cite{wang2018exploring}  &   -       &   92.2    &   -       &   91.2    &   -       &   1       &   50      \\
                MetaPruning \cite{Metapruning}  &   76.6    &   -       &   75.4    &   -       &   1.2     &   -       &   51.10   \\
                Autopr \cite{luo2018autopruner}&    76.15   &   92.87   &   74.76   &   92.15   &   1.39    &   0.72    &   51.21   \\
                %&GDP \cite{lin2018accelerating}    &   75.13   &   92.30   &   71.89   &   90.71   &   3.24    &   1.59    &   51.30   \\
                FPGM \cite{FPGM}                &   76.15   &   92.87   &   74.83   &   92.32   &   1.32    &   0.55    &   53.5    \\
                %&SSR-L2 \cite{lin2019towards}  &   75.12   &   92.30   &   71.47   &   90.19   &   3.65    &   2.11    &   55.76   \\  
                DCP \cite{zhuang2018discrimination}&76.01   &   92.93   &   74.95   &   92.32   &   1.06    &   0.61    &   55.76   \\      
                %&C-SGD \cite{CSGD}             &   75.33   &   92.56   &   74.54   &   92.09   &   0.79    &   0.47    &   55.76   \\
                ThiNet \cite{DBLP:journals/pami/LuoZZXWL19}&    75.30   &   92.20   &   72.03   &   90.99   &   3.27    &   1.21    &   55.83   \\
                EagleEye\textbf{*}$^1$ \cite{Li2020EagleEyeFS}& 77.21   &   93.68   &   76.37   &   92.89   &   0.84    &   0.79    &   50  \\
                %&SASL \cite{SASL}              &   76.15   &   92.87   &   75.15   &   92.47   &   1.00    &   0.40    &   56.10   \\
                \textbf{MaskSparsity (ours)}    &   \textbf{76.44}  &   \textbf{93.22}  &   \textbf{75.68}  &   \textbf{92.78}  &   \textbf{0.76}   &   \textbf{0.44}   &   \textbf{51.07}  \\
                %&TRP \cite{TRP}                    &   75.90   &   92.70   &   72.69   &   91.41   &   3.21    &   1.29    &   56.52   \\
                %&LFPC \cite{he2020learning}        &   76.15   &   92.87   &   74.46   &   92.32   &   1.69    &   0.55    &   60.8    \\
                
                %&DMC \cite{}               &   76.15   &   92.87   &   -   &   -   &   0.8 &   0.38    &   55.0    \\

                HRank \cite{HRank}     &    76.15   &   92.87   &   71.98   &   91.01   &   4.17    &   1.86    &   62.10   \\
                %&GAL-1-joint                   &   76.15   &   92.87   &   69.31   &   89.12   &   x       &   x       &   72.86   \\
                %&EagleEye \cite{}                  &   -   &   -       &   76.4    &   92.89       &   -       &   -       &   50  \\
                %   &HRank  \cite{HRank}                            &   76.15   &   92.87   &   69.10   &   89.58   &   7.05    &   3.29    &   76.03   \\
                \bottomrule[1.5pt]
                
            \end{tabular}
            \begin{tablenotes}
             \item[1] The baseline of EagleEye\textbf{*} is obtained by evaluating the weight provided by the authors.
            \end{tablenotes}
        \end{small}
        
        \end{threeparttable} 
    \end{center}
    \vspace{-0.7cm}
    \label{exp-table-imgnet}
\end{table*}
\section{Experiments}\label{Experiment}
\subsection{Experimental Settings}

\textbf{Datasets}.
To demonstrate the effectiveness of Mask Sparsity in reducing model complexity, we evaluate MaskSparsity on both small and large datasets,  \emph{i.e.}, CIFAR-10 \cite{krizhevsky2009learning} ,and ILSVRC-2012 \cite{russakovsky2015imagenet}. The CIFAR-10 dataset consists of natural images of 10 classes with resolution 32$\times$32 and the train and test sets contain 50,000 and 10,000 images respectively. The ImageNet dataset consists natural images of 1000 classes with resolution 224$\times$224 and the train and test sets contain 1.2 million and 50,000 images respectively. We experiment with ResNet-50 \cite{He2016IdentityMI} on ILSVRC-2012, and experiment with ResNet-56 \cite{he2016deep} and ResNet-110 \cite{he2016deep} on CIFAR-10.

\textbf{Codebase and Baseline}.
We directly deploy our algorithm on two popular codebases in github\footnote{https://github.com/weiaicunzai/pytorch-cifar100}$^,$\footnote{https://github.com/facebookresearch/pycls} for the experiments on cifar-10, cifar-100, and ILSVRC-2012. We hardly ever change any of the origin codes, except for adding the codes of our algorithm. 
Due to the difference in the training strategy of the baseline with other methods, we have the higher baseline accuracy on the evaluation datasets. It should be pointed that a higher baseline makes it difficult for the pruning algorithms to keep the accuracy after pruning.

%for the model to maintain a higher accuracy after pruning.

\textbf{Evaluation Protocols}.
We use the number of parameters and the FLOPS \cite{he2016deep} to evaluate the complexity of the networks. To evaluate the accuracy, we use top-1 and top-5 score of full-size models and pruned models on ILSVRC-2012 and top-1 score only on CIFAR-10.

\textbf{Training and Pruning setting}.
All the training-related hyper-parameters follow the two above-mentioned GitHub repositories. We use the same hyper-parameters during the normal training stage, the two sparsity training stages, and the finetuning stage in Figure \ref{MAskSparsityPipeline}. Specifically, on CIFAR-10, we train models for 200 epochs with a batch size of 128, a weight decay of 0.0005, a Nesterov momentum of 0.9 without dampening in every stage, and an initial learning rate of 0.1 which is divided by 5 at epochs 60, 120 and 160 on four NVIDIA GTX 1080Ti GPUs; On ILSVRC-2012, we train models for 100 epochs with a batch size of 256, a weight decay of 0.0001, a Nesterov momentum of 0.9 with dampening in every stage, and an initial learning rate of 0.1 which is divided by 10 at epochs 30, 60 and 90 on eight GPUs.

We set $\lambda$ as $2e^{-4}$ and $5e^{-4}$ separately for the global sparsity training stage and the mask sparsity training stage. We only manually set them without too much tuning.
We think the former should be set lower since it affects all channels of each layer. Moreover, in the two sparsity training stages, we reinitialize the network with a normally-trained model.
%and perform the mask sparsity training for the model by pruning the mask.

%To generate pruning masks, we first perform global sparsity training on the normally-trained networks. At this stage we use 

%with penalty factors $\lambda=2e^{-4}$, resnet110 with penalty factors $\lambda=5e^{-5}$ on CIFAR-10, and resnet50 with penalty factors $\lambda=2e^{-4}$ on ILSVRC-2012. Then we prune the sparse model based on the “smaller-norm-less-important” criterion and obtain the sparse mask according to the target pruning rate. 

%and a higher $\lambda$. On CIFAR-10, we choose $\lambda=1e^{-3}$ for ResNet-56 and ResNet-110, and we choose $\lambda=5e^{-4}$ for ResNet-50 On ILSVRC-2012. 

For the pruning mask generation step after the global sparse regularization stage, we use a threshold of $1e^{-2}$. This thresholding step is to generate a sparse mask for the mask sparse regularization stage. Also, we use this sparse mask as our final pruning mask after the mask sparse regularization. In this way, we perform the \emph{pruning-aware} sparse regularization on the network.
%The latter thresholding is for the final pruning and fine-tuning.

After pruning, we fine-tune the pruned models with an initial learning rate of 0.001, and keep other parameter settings the same as the previous step, on both the datasets.

%--------------------------------------------------------------------------------------------------------------------------------------------------

\subsection{Results and Analysis}

\subsubsection{Results on ILSVRC-2012}\label{results_on_ILSVRC-2012}
As shown in Table \ref{exp-table-imgnet}, our proposed MaskSparsity achieves the new state-of-the-art. NS \cite{liu2017learning} is the baseline method of MaskSparsity, which adopts the global sparse regularization on the scaling factors of BN layers. OT \cite{OT} is the improved NS method, which sets an optimal threshold for each layer. It can be seen that the MaskSparsity outperforms them significantly in the respect of accuracy drop (Top-1 $\downarrow$  and Top-5 $\downarrow$  in Table \ref{exp-table-imgnet}) under roughly the same level of FLOPS drop. This shows that with the fine-grained sparse regularization, we avoid the bad effect of the sparse regularization on the unpruned channels. Moreover, as shown in the ablation study \ref{abalation}, with MaskSparsity, the pruning threshold on the scaling factors is easier to set.

In Table \ref{exp-table-imgnet}, it can be seen that we also outperform the non-sparsity-training-based methods under the same FLOPS decrease rate, \eg, FPGM \cite{FPGM} (53.5\% FLOPS reduced), DCP \cite{zhuang2018discrimination} (55.76\% FLOPS reduced), MetaPruning \cite{Metapruning} (51.10\% FLOPS reduced), and EagleEye \cite{Li2020EagleEyeFS} (50\% FLOPS reduced). While the FLOPS reduction of MaskSparsity is less than HRank (53.76\% vs 62.1\%), the accuracy of the pruned model is much higher than that of HRank (0.93 vs 4.17 in Top-1 accuracy drop).
This shows MaskSparsity's superiority over the previous state-of-the-art methods.
\subsubsection{Results on CIFAR-10}\label{results_on_cifar10}

Table \ref{exp-table-cifar10-r56} shows the experimental results of ResNet-56 on CIFAR-10. On this small dataset, MaskSparsity also achieves the state-of-the-art performance. Under similar FLOPs reduction with FPGM \cite{FPGM} and Hinge \cite{Hinge}, MaskSparsity achieves $0.31\%$ Top-1 accuracy drop with ResNet-56, which is slightly better than FPGM \cite{FPGM} (0.31\% vs 0.33\%) and Hinge \cite{Hinge} (0.31\% vs 0.74\%).

Table \ref{exp-table-cifar10-r110} shows the experimental results of another network, ResNet-110 on CIFAR-10. With this deeper network, MaskSparity achieves a better performance. As shown in Table \ref{exp-table-cifar10-r110}, our MaskSparsity outperforms the other state-of-the-art methods, like HRank \cite{HRank} (at 58.2\% FLOPS reduction), under roughly the same ratio of FLOPS reduction. MaskSparsity has roughly the same accuracy increase with FPGM \cite{FPGM} (0.02 vs 0.06 )
%and C-SGD \cite{CSGD} (0.02 vs 0.06 and 0.03)
, but MaskSparsity has a larger FLOPS reduction than these two methods(63.03\% vs 52.3\% and 60.89\%).

%--------------------------------------------------------------------------------------------------------------------------------------------------

%--------------------------------------------------------------------------------------------------------------------------------------------------
\begin{table}[t]\small
    \small\caption{Evaluation results using ResNet-56 on CIFAR-10.}
    \vspace{-0.5cm}
    \setlength{\tabcolsep}{0.2em}
    \begin{center}
        \begin{tabular}{l|cccc} 
%           \hline
            \toprule[1.5pt]
            
            Method                          &\makecell[c]{Base \\ Top-1(\%)}        &\makecell[c]{Pruned \\ Top-1 (\%)} & \makecell[c]{Top-1 $\downarrow$ \\ (\%)}  &\makecell[c]{FLOPs$\downarrow$ \\ (\%)}                \\
%           \hline
            \midrule[1pt]
            
            %Li et al. \cite{li2016pruning}                 &   93.04   &   93.06           &   -0.02   &   27.60   \\
            %NISP-56 \cite{yu2018nisp}              &   -       &   -               &   0.03    &   43.61   \\
            %Zhu et al. \cite{zhu2018improving}                 &   93.39   &   93.40           &   -0.01   &   47.36   \\
            %L1 \cite{li2017pruning}    &93.26                            &93.06              &0.2   &14.1\% \\
            %
            
            %NISP \cite{yu2018nisp}   &93.26         &93.01   &0.025            &35.5\%       \\
            %
            
            %
            NISP \cite{yu2018nisp}   &-         &-   &0.03 &42.6 \\
            
            Hinge \cite{Hinge}   &93.69        &92.95   &0.74            &50       \\
            %CP \cite{he2017channel}    &   92.8    &   91.8            &   1.0     &   50      \\
            AMC \cite{he2018amc}                    &   92.8    &   91.9            &   0.9     &   50      \\
            %HRank \cite{HRank}                     &   93.26   &   93.17           &   0.09    &   50.0    \\
            %He \emph{et al}. \cite{he2017channel}    &93.26      &90.80   &2.46             &50.6       \\
            
            LeGR \cite{LeGR}   &93.9        &93.7   &0.2            &52 \\

            FPGM \cite{FPGM}                        &   93.59   &93.26  &   0.33    &   52.6\\
            %SFP \cite{he2018soft}                  &   93.59   &93.35  &   0.24    &   52.6\\
            LFPC \cite{he2020learning}              &   93.59   &93.24  &   0.35    &   52.9\\
            \textbf{MaskSparsity (ours)}            &   \textbf{94.50}  &   \textbf{94.19}  &\textbf{0.31}  & \textbf{54.88}    \\  
            %SASL \cite{SASL}                       &   93.63   &   93.58           &   0.05    &   57.1    \\
            GAL-0.8 \cite{GAL}   &93.26      &90.36     &2.9         &60.2  \\
            
            %GAL-0.8                                    &   93.26   &   91.58           &   1.68    &   60.2    \\
            %AFP \cite{ding2018auto}                    &   93.93   &   92.94           &   0.99    &   60.85   \\
            %C-SGD \cite{CSGD}                      &   93.39   &   93.62           &   -0.23   &   60.85   \\
            HRank \cite{HRank}      &   93.26   &   90.72           &   2.54    &   74.1    \\
            %ResRep (ours)                          &   93.71   &   92.94           &   0.67    &   75.12   \\
            %TRP \cite{TRP}                         &   93.14   &   91.62           &   1.52    &   77.82   \\
            
%           \hline
            \bottomrule[1.5pt]
%       \vspace{-0.8cm} 
        \end{tabular}
        \vspace{-0.4cm} 
    \end{center}
    \label{exp-table-cifar10-r56}
\end{table}
%--------------------------------------------------------------------------------------------------------------------------------------------------

%--------------------------------------------------------------------------------------------------------------------------------------------------
\begin{table}[t]
    \large
%   \vspace{-0.4cm}
    \small
    \setlength{\tabcolsep}{0.2em}
    \begin{center}\caption{Evaluation results using ResNet-110 on CIFAR-10.}    
    \vspace{-0.4cm}
    \label{exp-table-cifar10-r110}
        \begin{tabular}{l|cccc}
%           \hline
            \toprule[1.5pt]
            
            Method                          &\makecell[c]{Base\\Top-1(\%)}      &\makecell[c]{Pruned\\Top-1(\%)}    & \makecell[c]{Top-1 $\downarrow$ \\ (\%)}  &\makecell[c]{FLOPs$\downarrow$\\(\%)}              \\
%           \hline
            \midrule[1pt]

            Li et al. \cite{li2016pruning}          &   93.53   &   93.30   &   0.23        &   38.60       \\
            SFP \cite{he2018soft}                   &   93.68   &   93.86   &   -0.18       &   40.8        \\
            NISP-110 \cite{yu2018nisp}              &   -       &   -       &   0.18        &   43.78       \\
            GAL-0.5 \cite{GAL}                      &   93.50   &   92.74   &   0.76        &   48.5        \\
            
            FPGM \cite{FPGM}        &93.68  &93.74  &   -0.06   &   52.3\\

            HRank \cite{HRank}                      &   93.50   &   93.36   &   0.14        &   58.2        \\
            
            LFPC \cite{he2020learning}              &   93.68   &   93.07   &   0.61        &   60.3        \\
            %C-SGD \cite{CSGD}                      &   94.38   &   94.41   &   -0.03       &   60.89       \\
            \textbf{MaskSparsity (ours)}    &   \textbf{94.70}  &   \textbf{94.72}  &   \textbf{-0.02}  &   \textbf{63.03}\\
            HRank \cite{HRank}      &   93.50   &   92.65   &      0.85     &   68.6        \\

            %           C-SGD \cite{CSGD}                       &   94.38   &   94.41   &   -0.03       &   60.89       \\
            %HRank                                  &   93.50   &   92.65   &   0.85        &   68.6        \\
            
            SASL \cite{SASL}                        &   93.83   &   93.80   &   0.03        &   70.2        \\

%           \hline
            \bottomrule[1.5pt]
%       \vspace{-0.5cm}     
        \end{tabular}
        \vspace{-0.5cm}
    \end{center}
    
\end{table}
%--------------------------------------------------------------------------------------------------------------------------------------------------

Table \ref{exp-table-cifar10-VGG16} shows the experimental results of VGG-16, which is a straight network structure that is different from ResNet. We compare MaskSparsity with NetSlim (NS) \cite{liu2017learning}, FPGM \cite{FPGM}, and PFEC \cite{li2016pruning}. It shows that we outperforms NS \cite{liu2017learning} and PFEC \cite{li2016pruning} on both accuracy and FLOPS reduction. Compared with FPGM \cite{FPGM}, we are 0.04 \% less than FPGM on the increase of the accuracy, which is very minor. However, we decrease 18.01\% more FLOPS than FPGM. Therefore, we also outperform FPGM in general pruning performance. Moreover, we also compare the accuracy drop of the pruned model without the fine-tuning stage. It can be seen in the third column of Table \ref{exp-table-cifar10-VGG16} that the MaskSparsity suffers a weaker accuracy drop than the other methods.

\begin{table}[t]
    \caption{Evaluation results using VGG-16 on CIFAR-10. FT: fine-tuning.}
    \vspace{-0.6cm}
%   \linespread{1.5}
    %\small
    %\scriptsize
    \footnotesize
    \setlength{\tabcolsep}{0.2em}
    \begin{center}
        \begin{tabular}{l|ccccc}
            \toprule[1.5pt]
            Method            % & \multicolumn{5}{c}{Pruned} \\ \cline{2-6}
            & \makecell[c]{Base\\Top-1(\%)}    & \makecell[c]{before\\FT(\%)} & \makecell[c]{FT \\ (\%)}  & \makecell[c]{Top-1 $\downarrow$ \\ (\%)} &\makecell[c]{FLOPs$\downarrow$ \\ (\%)} \\ \midrule[1pt]
%           Method          &Base Top-1     &\shortstack {Pruned\\before FT} &FT    & Top-1 $\downarrow$\%  &FLOPs$\downarrow$\% \\ 
%           \hline
            
%           PFEC~\cite{li2016pruning} {\scriptsize{( from \cite{FPGM})}}   & 93.58  &77.45 & 93.28 & 0.3  &34.2\\  
            
            PFEC~\cite{li2016pruning}  & 93.58  &77.45 & 93.28 & 0.3  &34.2\\  
            
            FPGM \cite{FPGM}   & 93.58  &80.38 & 94.00  & {-0.42} &34.2\\  
            
            NS \cite{liu2017learning} & 93.66 &-  & 93.80  &-0.14 &51\\  
            
            \textbf{MaskSparity(ours)}                &\textbf{93.86}   &\textbf{94.16} &\textbf{94.24} &\textbf{-0.38} & \textbf{52.21}\\ 
            
            \bottomrule[1.5pt]
            
%           \hline
        \end{tabular}
                    \vspace{-0.5cm}

    \end{center}
    \label{exp-table-cifar10-VGG16}
\end{table}
%--------------------------------------------------------------------------------------------------------------------------------------------------

\begin{figure}[t]
    \begin{centering}
        \includegraphics[width=0.4\textwidth]{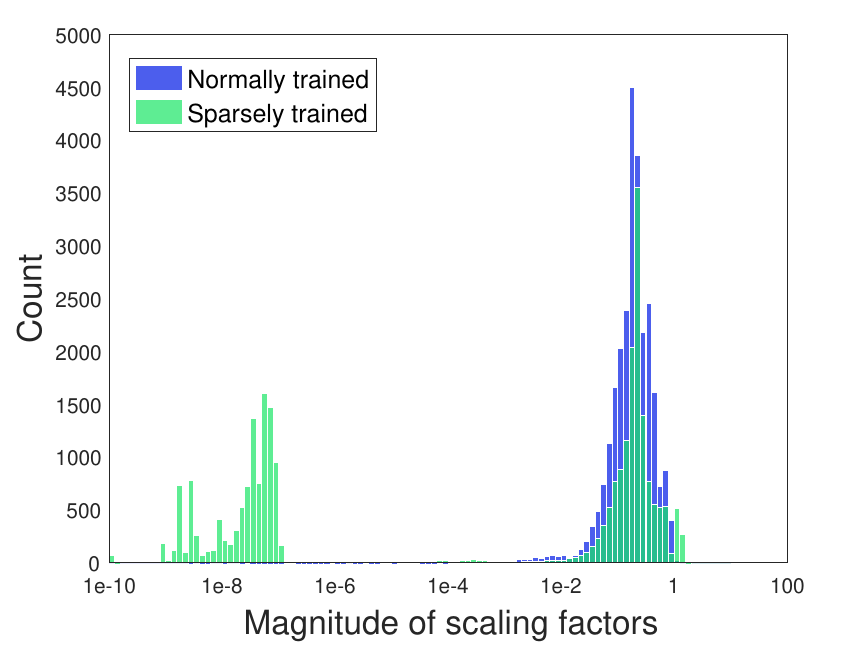} 
        \vspace{-0.4cm}
        \caption{Distribution of scaling factors of ResNet50 on ILSVRC-2012 before and after the MaskSparsity's sparsity training.}
        \label{dstbtion-mask-sp}
                \vspace{-0.6cm}
        \par\end{centering}
\end{figure}

\begin{figure}[t]
    \begin{centering}
        \includegraphics[width=0.42\textwidth]{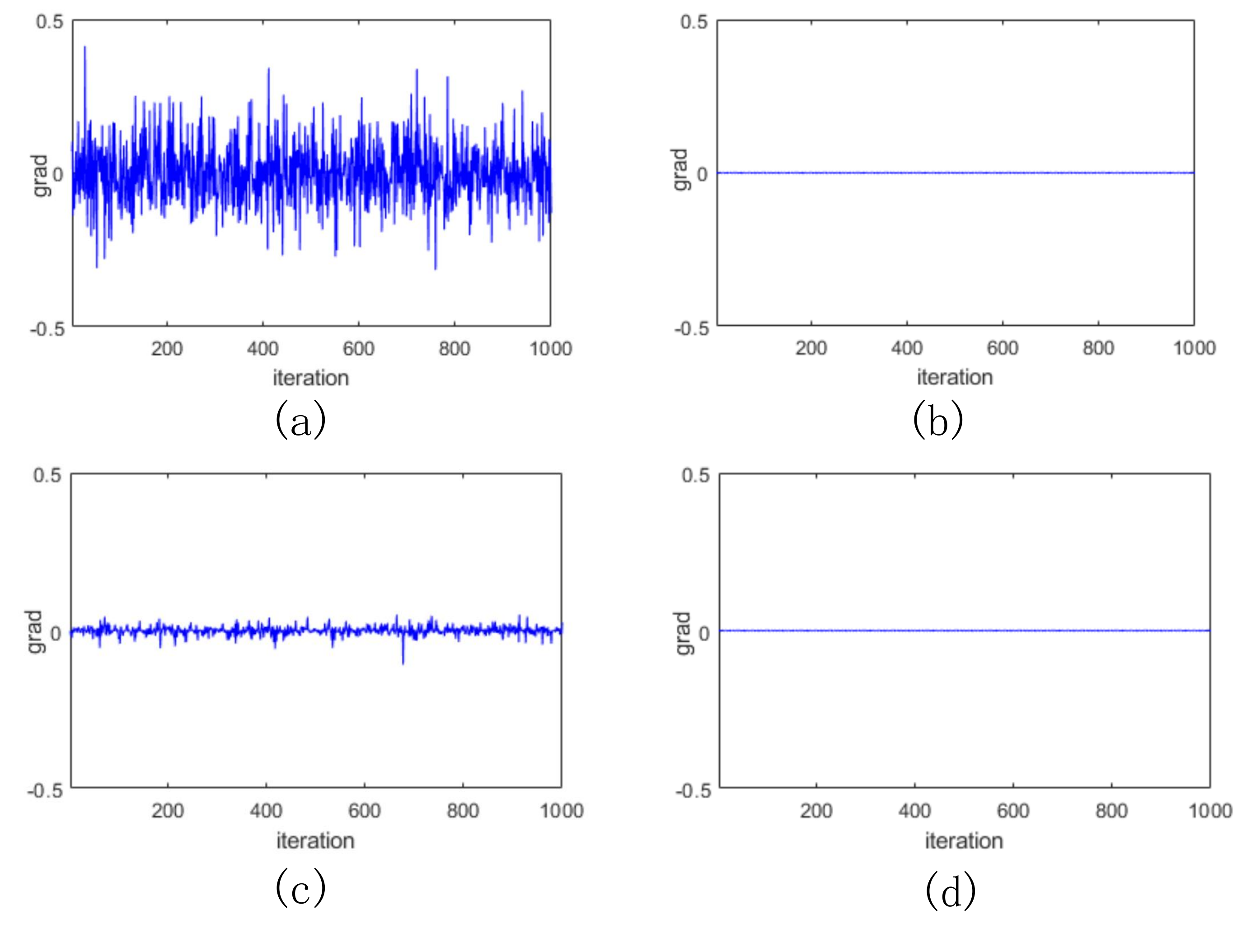}  
        \vspace{-0.3cm}
        \caption{The gradients of the scaling factors of some certain channel at the end of the sparsity training stage. Both important and unimportant channels are visualized. (a) The important channel, global sparsity. (b) The unimportant channel, global sparsity. (c) The important channel, MaskSparsity. (d) The unimportant channel, MaskSparsity.}
                \vspace{-0.2cm}
        \label{grad}
        \par\end{centering}
    
\end{figure}

\subsection{Ablation Study}\label{abalation}
\textbf{Visualization of the distribution of the scaling factors after using MaskSparsity.} 
In Section \ref{Analysisoftraditionalsp} and Figure \ref{pbm}, we show that the distribution of scaling factors meets the over-regularization problem that might damage the pruning performance. In Figure \ref{dstbtion-mask-sp} we draw the distribution of the same set of scaling factors after the mask-guided sparse regularization and compare it with that of the pre-trained networks. It can be seen that the two problems are alleviated significantly. The right peak of the sparsity trained network does not move towards 0 like in Figure \ref{pbm}. Moreover, the two peaks are well distinguishable, with almost no in-between bars in the middle area. This demonstrates the effectiveness of the fine-grained sparse regularization of MaskSparsity, which would benefit the pruning performance.

\textbf{The convergence analysis by visualizing the gradients.} To validate the statement that the sparse regularization on unpruned channels restricts the network's capacity, we visualize the gradients of important and unimportant channels after the sparsity training using global and fine-grained sparse regularization, respectively. The result is shown in Figure \ref{grad}. The gradients are collected by continuing training for 1,000 iterations from the end of the sparsity training stage of ResNet-50 on ILSVRC-2012. In Figure \ref{grad} (a), it can be seen that the gradient norm of the important channels is still large for the important channels with the global sparse regularization, while Figure \ref{grad} (b) shows the important channel has a small gradient norm with our fine-grained MaskSparsity sparse regularization. Figure \ref{grad} (b) and (d) shows that both the gradients of unimportant channels with the two kinds of sparse regularization methods are almost the same. According to these figures, although the network's accuracy and sparsity have converged, the global sparse regularization leads to a larger gradient on the important channels. We infer that the large gradient prevents the network converges to a better local minimum.

\textbf{Comparing fine-tuning and train-from-scratch.} 
In Table \ref{Finetune_scratch_cifar10}, we list the network model's accuracy and computational complexity at different pruning stages. Moreover, we also list the result that trains the pruned model from scratch without exploiting its weights. It can be seen that the train-from-scratch result is lower than the fine-tuning result. We think this demonstrates the effectiveness of the fine-grained sparsely-trained pre-trained weights.

% \textbf{The robustness to various pruning thresholds.} Table \ref{pruning_threshold} shows the accuracy of the pruned model under different pruning thresholds. It can be seen that the accuracy and model complexity of the pruned model change only a little in the range of $[1e^{-4}:1e^{-2}]$. This shows that the MaskSparsity is robust to many threshold settings and it does not need a carefully selected threshold. 

\begin{table}[t]
%   \vspace{-0.2cm}
    \caption{The stage-wise performance in the case of  pruning ResNet-56 on CIFAR-10.}
    \vspace{-0.4cm}
    \setlength{\tabcolsep}{0.5em}
    \begin{center}\small
        \begin{tabular}{l|ccc}
            \toprule[1.5pt]
            Model state               &Top-1(\%)           &FLOPs          &Parameters \\
            \midrule[1pt]
            Normally trained      &94.50            &126M      &853K  \\
            MaskSparsity trained        &94.02             &126M      &853K  \\

            Pruned      &92.67             &57M     &419K  \\
            Finetune       &94.19             &57M    &419K \\
            \hline
            Train from Scratch   &93.60            &57M  &419K  \\
            \bottomrule[1.5pt]
            
        \end{tabular}
    \end{center}    \vspace{-0.6cm}
    \label{Finetune_scratch_cifar10}
\end{table}

\textbf{The performance on the pruning mask generators.}
As discussed above, the above pruning process consists of two key elements, i.e. identifying the unimportant channels and pushing them to 0 by sparse regularization. This paper uses global sparsity training to generate the pruning mask. To show the MaskSparsity's generalization on other pruning mask generators, we apply it to two other pruning methods and show the superiority on the performance. Except for the pruning mask generating method, the other stage is the same as the pipeline in Figure \ref{MAskSparsityPipeline}.

Firstly, we directly use the codebase of EagleEye \cite{Li2020EagleEyeFS} and use the pruning mask after its searching process to conduct our mask sparse regularization stage. We reuse the hyper-parameters of EagleEye's finetuning stage in our sparse regularization. Table \ref{Masksparsity EagleEye} shows the experimental results. Under the same pruning mask after pruning, the top-1 accuracy
increased by 0.32\% over the original EagleEye. This
shows the scalability of MaskSparsity on state-of-the-art methods.

Secondly, we try the naive pruning method that directly pruning the same portion of channels of each layer. We call this naive pruning method as \emph{Uniform Pruning}. Experimental results on ResNet-50 are shown in Table \ref{Masksparsity vs Direct pruning}.  It can be seen that MaskSparsity improves this naive pruning method to SOTA-level performance. This demonstrates the effectiveness of MaskSparsity on pushing the unimportant channels to near 0 without too much damage on the important channels.

%of final pruned resnet50 on the Imagenet dataset using different masks in masksparsity stage. 
%Slim mask is obtained by step 3 of Algorithm \ref{alg:MaskSP}; Contrary Slim Mask is obtained by selecting the part channels with the largest scaling factors to prune and maintaining the same channel number configuration of each layer with Slim Mask; Uniform Mask is obtained by pruning each layer with the same pruning rate and maintaining closed FLOPs as model pruned by Slim Mask. 
%It can be seen that the result of using the Slim Mask has the highest accuracy, the Uniform Mask the second, and the Contrary Slim Mask the smallest. There is little difference among the three results. Even with a rough mask, our method can still achieve good results.  

\begin{table}[t]
%   \vspace{-0.2cm}
    \caption{Evaluation results of MaskSparsity using the pruning mask of EagleEye.}
    \vspace{-0.4cm}
    \setlength{\tabcolsep}{0.5em}
    \begin{center}\small
        \begin{tabular}{l|ccc}
            \toprule[1.5pt]
            Model state              &Top-1(\%)           &Top-5(\%)            &FLOPs$\downarrow$(\%)   \\
            \midrule[1pt]
            Unpruned Resnet-50                  &77.21             &93.68              & -  \\
            EagleEye\cite{Li2020EagleEyeFS}                 &76.37             &92.89              &50.0  \\
            MaskSparsity + EagleEye  &76.69             &93.22              &50.0  \\
            \bottomrule[1.5pt]
            
        \end{tabular}
            \vspace{-0.3cm}
    \end{center} 
    \label{Masksparsity EagleEye}
\end{table}

\begin{table}[htbp]
  % \vspace{-0.2cm}
    \caption{Evaluation results of MaskSparsity based on uniform pruning mask of ResNet-50 on ImageNet.}
    \vspace{-0.4cm}
    \setlength{\tabcolsep}{0.4em}
    \begin{center}\small
        \begin{tabular}{l|ccc}
            \toprule[1.5pt]
            Model state                          &\makecell[c]{Top-1 \\ (\%)}           &\makecell[c]{Top-5 \\(\%)}            &\makecell[c]{FLOPs$\downarrow$ \\ (\%)} \\
            \midrule[1pt]
            Unpruned Resnet-50                            &76.44             &93.22              & -  \\
            Direct pruning with Uniform Mask    &74.23             &92.28              &53.46  \\
            MaskSparsity with Uniform Mask       &75.62             &92.68              &53.46  \\

            \bottomrule[1.5pt]
            
        \end{tabular}
            \vspace{-0.6cm}
    \end{center} 
    \label{Masksparsity vs Direct pruning}
\end{table}

\begin{table}[htbp]

    \caption{Evaluation results of MaskSparsity with different regularizations of ResNet-56 on CIFAR-10.}   \vspace{-0.4cm}
%   \vspace{-0.3cm}
    \setlength{\tabcolsep}{0.5em}
    \begin{center}\small
        \begin{tabular}{l|ccc}
            \toprule[1.5pt]
            Model state                        &Top-1(\%)           &Top-5(\%)            &FLOPs$\downarrow$(\%) \\
            \midrule[1pt]
            Unpruned Resnet-56                            &94.50             &99.79              & -  \\
            MaskSparsity with L1               &94.19             &99.81              &54.88  \\
            MaskSparsity with L2               &93.99             &99.8               &54.88  \\
            \bottomrule[1.5pt]
            
        \end{tabular}
        
            \vspace{-0.5cm}
            
    \end{center} 
    \label{Masksparsity with different Norm}
\end{table}

\textbf{Masksparsity with different Regularization.} 
In this paper, we mainly use the L1 regularization to generate the network sparsity. To show the compatibility of MaskSparsity with other specific forms of regularizations,  we replace the L1 regularization with L2 regularization in the mask sparse training stage. We conduct this ablation study using ResNet-56 on CIFAR-10. The experimental results are shown in Table \ref{Masksparsity with different Norm}. It can be seen that there is little difference in accuracy between the result of L1 regularization in MaskSparsity stage and L2 regularization. The accuracy of L1 regularization in MaskSparsity stage is 0.2 points higher than L2 regularization.

\subsection{Application on Other Tasks}

To validate the generalization ability, We apply the method to two different object detection tasks. The first is the face detection based on YOLOv5\footnote{https://github.com/ultralytics/yolov5} evaluated on WiderFace\cite{yang2016wider}. The second is the car detection based on Faster-RCNN-FPN \cite{lin2017feature} evaluated on PASCAL VOC. The results are listed in Table \ref{Masksparsityyolov5} and Table \ref{Masksparsityvoc}. For Faster-RCNN-FPN, we only prune the backbone part and report the backbone's FLOPS and parameters. From these experimental results, the pruned models of both tasks maintain roughly the same level of accuracy. It can be concluded that MaskSparsity applies to other tasks.

\begin{table}[t]

    \caption{Evaluation results of MaskSparsity on an YOLOv5s-based face detector on WiderFace.}    \vspace{-0.4cm}
%   \vspace{-0.3cm}
    \setlength{\tabcolsep}{0.5em}
    \begin{center}\small
        \begin{tabular}{l|ccc}
            \toprule[1.5pt]
            Model state                        & mAP[Easy]        &   FLOPS           &Param \\
            \midrule[1pt]
            YOLOv5s                            &92.38\%             &4.1G              & 3.5M  \\
            YOLOv5s+MaskSparsity               &91.86\%             &2.0G              &1.6M  \\
            \bottomrule[1.5pt]
            
        \end{tabular}
        
            \vspace{-0.5cm}
            
    \end{center} 
    \label{Masksparsityyolov5}
\end{table}

\begin{table}[t]
    \caption{Evaluation results of MaskSparsity on an FPN-based car detector on PASCAL VOC. The input size is 1000$\times$600 and the backbone is ResNet-50.}   \vspace{-0.4cm}
%   \vspace{-0.3cm}
    \setlength{\tabcolsep}{0.5em}
    \begin{center}\small
        \begin{tabular}{l|ccc}
            \toprule[1.5pt]
            Model state                        & mAP        &   FLOPS           &Param \\
            \midrule[1pt]
            Faster-RCNN-FPN                            &89.7\%             &49.95G              & 23.5M  \\
            Faster-RCNN-FPN+MaskSparsity               &89.3\%             &23.73G              &11.4M  \\
            \bottomrule[1.5pt]
        \end{tabular}
            \vspace{-0.5cm}
    \end{center}
    \label{Masksparsityvoc}
\end{table}

\section{Conclusion}\label{Conclusion}

In this paper, to solve the problem that existing sparsity-training-based methods over-regularize the important channels, we design a pruning-aware sparse training method,  named as MaskSparsity. MaskSparsity only applies the sparse regularization on the unimportant channels which are to be pruned. Therefore, MaskSparsity can minimize the negative impact of the sparse regularization on the important channels. The method is effective and efficient. The experimental results show that it outperforms the other sparsity-training-based pruning methods and achieves the state-of-the-art on the benchmarks. In the future, we plan to work on how to obtain better pruning masks.

%%%%%%%%% REFERENCES
{\small
\bibliographystyle{ieee_fullname}
\bibliography{egbib}
}

\end{document}